\documentclass[conference]{IEEEtran}
\IEEEoverridecommandlockouts
\usepackage{cite}
\usepackage{amsmath,amssymb,amsfonts}
\usepackage{algorithmic}
\usepackage{graphicx}
\usepackage{textcomp}
\usepackage{xcolor}

\usepackage{array}
\usepackage{colortbl}
\usepackage{multirow}
\usepackage{booktabs} 
\usepackage{diagbox} 
\usepackage{adjustbox}
\usepackage{pifont}
\usepackage{wrapfig}
\usepackage{floatflt}

\def\BibTeX{{\rm B\kern-.05em{\sc i\kern-.025em b}\kern-.08em
    T\kern-.1667em\lower.7ex\hbox{E}\kern-.125emX}}
\begin{document}

\title{A Comprehensive Review of Multimodal Facial State Analysis: Tasks, Methods, and Resources}
\author{Xuri Ge$^1$, Tianshuo Zhang$^1$, Ruihan Li$^1$, Hui Ye$^{1*}$$\thanks{* Corresponding author.}$, \\ Kaiwen Zheng$^{2*}$, Junchen Fu$^2$, Da Huo$^2$, Joemon M. Jose$^2$, Hu Han$^3$
\\ 
    {\normalsize
    $^{1}$Shandong University, Qingdao Key Laboratory of Trustworthy Artificial Intelligence, Shandong, China}\\
    {\normalsize
    $^{2}$University of Glasgow, United Kingdom}
    {\normalsize
    $^{3}$Institute of Computing Technology, Chinese Academy of Sciences, Beijing, China} \\
    {\normalsize
    \texttt{xuri.ge@sdu.edu.cn, \{tianshuozhang,ruihanli,huiyee\}@mail.sdu.edu.cn,}}\\
    {\normalsize
    \texttt{\{k.zheng.1, j.fu.3\}@research.gla.ac.uk,joemon.jose@glasgow.ac.uk, hanhu@ict.ac.cn}}
}

\maketitle

\begin{abstract}
Facial state analysis plays a crucial role in understanding human expressions, psychological modeling, and human–computer interaction. Traditional unimodal vision-based methods are often limited by environmental sensitivity and weak interpretability. Multimodal facial state analysis addresses these issues by integrating complementary cues from visual, audio, textual, physiological, and other related modalities. This survey emphasizes two key aspects: on one hand, multimodal learning enables contextual semantic understanding for improved facial state reasoning and leverages interpretable language generation to enhance model explainability; on the other hand, multi-task learning allows simultaneous analysis of expressions, action units (AUs), and face-based soft biometrics (e.g., age, gender), effectively capturing fine-grained expressions and improving cross-scene generalization. This survey reviews core tasks, representative methods, and datasets in multimodal facial state analysis, focusing on facial expression recognition, AU detection, and face-based soft biometric estimation, and emphasizing the unique value of language in providing contextual semantics, enhancing reasoning, and generating explanations. The survey aims to provide an up-to-date overview of the literature and to highlight future research directions for multimodal, interpretable, and multi-task adaptive facial state analysis.
\end{abstract}


\section{Introduction}
\label{sec:intro}
Facial state analysis plays a crucial role in understanding human expressions, psychological modeling, and human–computer interaction. As early as the 19\textsuperscript{th} century, pioneering studies~\cite{duchenne1876mecanisme, darwin_expression} demonstrated that facial muscle movements can precisely convey internal states. 
Subsequent research in psychology~\cite{ekman1992argument} has further confirmed that facial muscle activities and appearance changes are among the most powerful and universally understood nonverbal cues in social interactions.
With the rapid advancement of computer vision techniques, analyzing facial states through automated computational methods has become an increasingly important research direction~\cite{zhao2016deep, khan2019unified}, providing scalable, objective, and real-time tools to understand human psychological states.
\begin{figure}[t]
    \centering
		\includegraphics[width=1.0\linewidth]{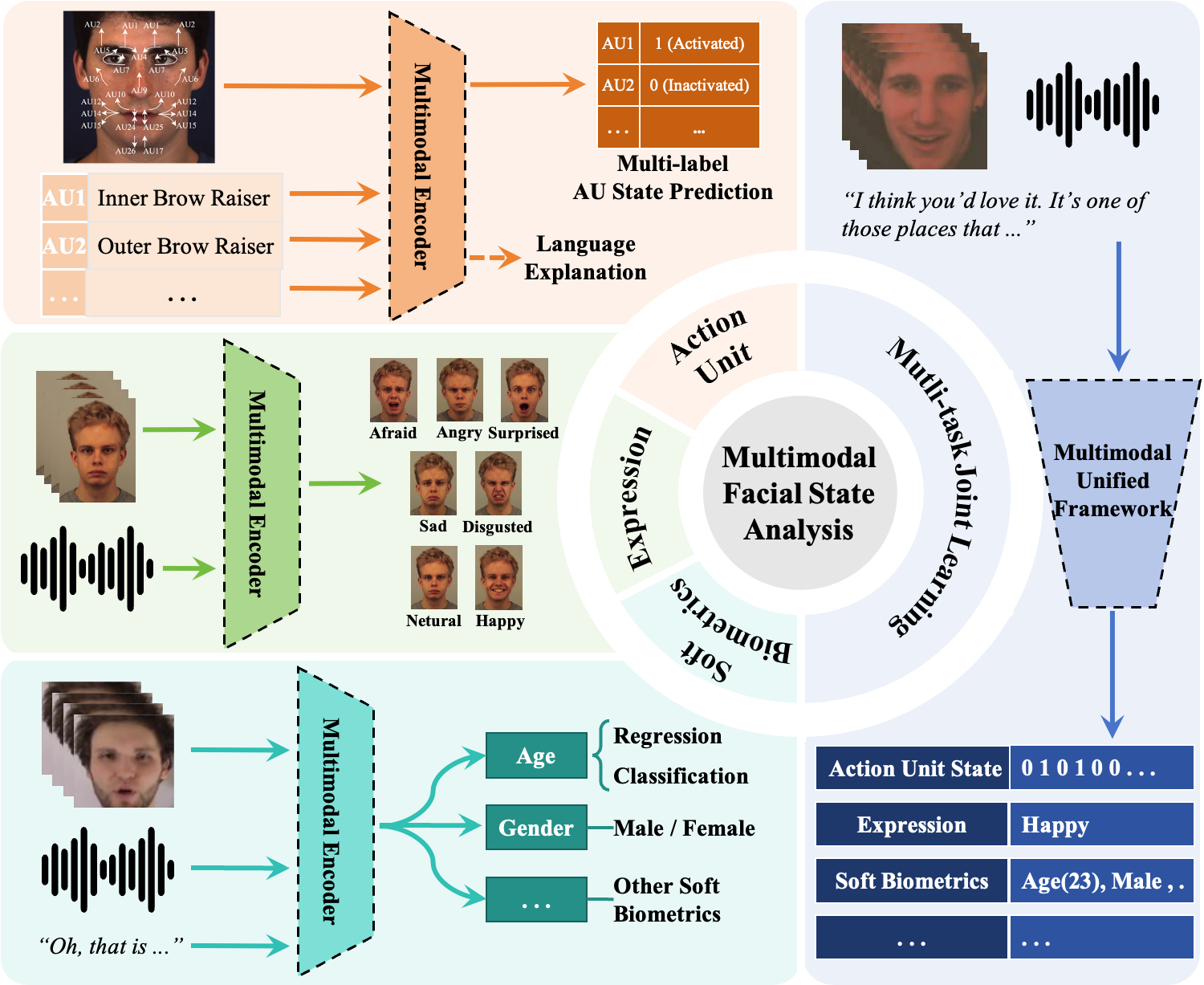}
        \vspace{-1.5em}
     \caption{Multimodal facial state analysis integrates visual information and other complementary modalities, providing a comprehensive assessment of facial action units (AU), expressions, and soft biometrics. By leveraging the intrinsic correlations among tasks, multi-task learning frameworks further strengthen feature robustness.
     }
    	\label{fig:overview}    
     	\vspace{-1.5em}
\end{figure}


Early facial state analysis methods were predominantly unimodal~\cite{mittal2020emoticon}, relying only on visual inputs, such as face images or videos. While achieving notable benchmark performance, these vision-based models remain vulnerable to environmental factors, such as lighting variations and occlusions, which can substantially impair their reliability in real-world applications.
Moreover, purely vision-based studies often produce only prediction scores, lacking clear interpretability aligned with human-understandable semantics, which limits their transparency and practical utility in sensitive contexts.

To overcome these limitations, researchers have increasingly turned to multimodal facial state analysis~\cite{yang2021exploiting, wang2023pose}, which integrates visual data with other modalities such as audio and text. By enabling interaction across modalities, these methods exploit complementary cues, mitigate the ambiguity and noise inherent in single-modality inputs, and generate robust representations that enhance task performance.
In addition, \cite{ge2024towards} has extended model outputs by generating textual explanations alongside the predicted results, thereby improving model interpretability and the trustworthiness of the results.

In this survey, we will limit our discussion to the three most commonly studied tasks in multimodal facial state analysis (see Fig.~\ref{fig:overview}), which together characterize complementary aspects of facial states:
(i) facial action unit (AU) recognition~\cite{chang2024facial}, decomposing facial muscle movements into anatomically grounded unit states~\cite{ekman1978facial} by integrating multimodal cues;
(ii) facial expression recognition (FER)~\cite{shi2024detail}, capturing transient affective states by jointly modeling visual and non-visual signals;
(iii) facial soft biometrics recognition~\cite{shou2023cilf}, which leverages multimodal information to complement visual appearance and identify relatively stable facial-based soft biometrics, including age, gender, and other attributes.

\begin{table*}[t]
    \centering
    \caption{Categorization of existing facial state databases. Task(s): AU=Facial action unit recognition, Exp=Facial expression recognition, Bio=Facial soft biometrics recognition. Modality: V=Visual, A=Audio, T=Textual, P=Physiological. Text: C=Caption, R=Reasoning, D=Description. Annotations: VA=Valence and arousal.}
    \small
    \setlength{\tabcolsep}{1pt}
    \fontsize{8}{9}\selectfont
    \renewcommand\tabcolsep{4pt}  
    \begin{adjustbox}{max width=\textwidth}
    \begin{tabular}{l|c|c|c|c|c|c}
        \toprule
        \textbf{Dataset} & \textbf{Task(s)} & \textbf{Modality} & \textbf{Text} & \textbf{Annotations} & \textbf{Subject Size} & \textbf{Samples} \\
        \midrule
        \multicolumn{7}{c}{\textbf{Single-task Multi-Modal Dataset}} \\
        \midrule
         BP4D+~\cite{zhang2016multimodal}     & AU   & V,P   & - & 34 AUs               & 140   & 1,400 sequences        \\
         MER2023 \cite{MER2023}               & Exp  & V,A   & - & 6 emotions, Valence  & -     & 3,373 videos  \\ 
         CMU-MOSEI~\cite{zadeh2018multimodal} & Exp  & V,A,T & C & 6 emotions           & 1,000 & 23,453 videos \\  
         DEAP~\cite{koelstra2011deap}         & Exp  & V,P   & - & VA                   & 32    & 1,280 sequences  \\  
         MERR~\cite{Emotion-llama}            & Exp  & V,T   & R & Instruction-response pairs & -     & 28,618 coarse, 4,487 fine pairs \\
         AgeVoxCeleb~\cite{tawara2021age}     & Bio  & V,A   & - & Age                  & 4,976 & 21,707 videos \\
         SpeakingFaces~\cite{SpeakingFaces}   & Bio  & V,A   & - & Age, Gender          & 142   & 13K instances    \\ 
        \midrule

        \multicolumn{7}{c}{\textbf{Multi-task Multi-Modal Dataset}} \\
        \midrule
        Aff-Wild2~\cite{kollias2019expression} & AU, Exp      & V,A & - & VA, 7 emotions, 8 AUs         & 558 & 2.8M frames (558 videos)      \\
        BU-EEG~\cite{li2020eeg}                & AU, Exp      & V,P & - & 6 emotions, 10 AUs            & 29  & 2,320 sequences                  \\
        FaceInstruct-1M \cite{face-llava}      & AU, Exp, Bio & V,T & R & Instruction-response pairs    & -   & 850K images, 120 hours video \\
        FABA-Instruct \cite{FABA}              & AU, Exp      & V,T & R & Instruction-response pairs    & -   & 30K pairs (19K images) \\
        FaceBench \cite{FaceBench}             & Exp, Bio     & V,T & D & Question-answer pairs         & -   & 74K pairs (16K images)     \\
        FEA-20K \cite{FEA-20K}                 & AU, Exp      & V,T & R & Instruction-response pairs    & -   & 19{,}425 pairs               \\  

        \bottomrule
    \end{tabular}
    \end{adjustbox}
    \label{tab:dataset_overview2}
    \vspace{-1em}
\end{table*}

Furthermore, these three tasks are deeply interconnected: AUs describe fine-grained muscle activations, which serve as the physical basis of facial expressions, while soft biometrics provide relatively stable structural priors that constrain and modulate both AU activations and expression patterns.
These intrinsic relationships enable multi-task learning~\cite{kollias2023multi}, which allows simultaneous modeling of AUs, expressions, and soft biometrics within a unified multimodal framework.

Despite increasing recognition of multimodal modeling and the intrinsic relationships among different facial dimensions, a comprehensive and unified review of multimodal facial state analysis remains lacking.
Existing surveys~\cite{martinez2017automatic, li2020deep} on facial state analysis largely concentrate on traditional unimodal settings. More recently, multimodal FER has been surveyed in \cite{ezzameli2023emotion}; nevertheless, expression is treated as an isolated problem, overlooking the inherent correlations among AU, expressions, and soft biometrics.
To address this gap, this survey systematically reviews core tasks, datasets, and representative methods in multimodal facial state analysis, and further discusses the importance of jointly learning these complementary tasks.

\section{Tasks}

Facial state analysis is inherently multidimensional, encompassing correlated tasks ranging from fine-grained muscle dynamics to long-term physiological traits.
This section introduces three core tasks: action unit (AU), expression, and soft biometrics recognition.
Building on the intrinsic correlations among these tasks, the section further presents multi-task joint learning in multimodal facial state analysis.

\subsection{Multimodal Facial Action Unit Recognition}
Facial action units (AUs) are fundamental components of facial expression, representing a specific anatomical facial muscle movement as defined by Ekman and Friesen~\cite{ekman1978facial}.
Typically, AU recognition is formulated as a multi-label binary classification task~\cite{shao2018deep}, aiming to detect the presence of specific AUs (e.g., Cheek Raiser AU6, Lip Corner Puller AU12).
Although early research primarily relied on visual cues from static images or dynamic videos~\cite{ge2021local}, recent work has increasingly shifted toward multimodal AU recognition~\cite{zhang2022transformer, zhang2023weakly}.
Multimodal AU recognition integrates complementary modalities, such as natural language descriptions~\cite{chang2024facial} and audio signals~\cite{yu2025towards}, to enrich visual cues, thereby yielding a robust and comprehensive representation of facial muscle movements.
This multimodal framework treats the task as a cross-modal representation problem, aiming to learn unified embeddings that fuse visual and non-visual signals to precisely characterize anatomically grounded AUs.
In addition to multimodal inputs, a growing line of work~\cite{ge2024towards} explores multimodal outputs within the AU recognition framework, where models leverage cross-modal reasoning to produce natural-language explanations alongside AU predictions, offering interpretability that traditional vision-only approaches cannot provide.

\subsection{Multimodal Facial Expression Recognition}
Unlike facial AU recognition, which analyzes fine-grained facial muscle activations, facial expression recognition (FER) focuses on inferring higher-level affective states from facial appearance.
Traditionally, FER has been treated as a unimodal visual task, relying solely on RGB images to infer emotional intent~\cite{mittal2020emoticon}. However, visual cues alone can be unreliable due to lighting changes, occlusions, and head-pose variations.
To overcome these limitations, recent studies have increasingly shifted toward multimodal FER, which jointly leverages complementary information from multiple modalities~\cite{zhang2023weakly}.
Grounded in psychological research, multimodal FER is generally divided into: (i) discrete emotion classification~\cite{wang2023pose}, which assigns expressions to predefined labels (e.g., happiness, surprise) following categorical emotion theory~\cite{ekman1992argument}; (ii) continuous dimensional regression~\cite{ahire2025maven}, which maps expressions onto a valence–arousal space according to dimensional emotion theory~\cite{russell1980circumplex}.
Regardless of the specific paradigm, multimodal FER relies on the synergy of heterogeneous signals, as human emotion is inherently multimodal—manifesting through facial expressions, vocal intonations, linguistic content, and physiological responses. Fusing these diverse modalities mitigates the ambiguity of visual-only analysis, ensuring robust and accurate emotion understanding in real-world scenarios.

\subsection{Multimodal Facial Soft Biometrics Recognition}
Facial soft biometrics recognition focuses on estimating non-unique but relatively stable facial attributes, such as age, gender, ethnicity, and facial shape~\cite{hassan2024soft}.
Beyond traditional vision-based methods~\cite{agbo2020deeply}, emerging research has begun exploring multimodal facial soft biometrics recognition.
In this paper, we limit our scope to the most commonly studied tasks: multimodal age estimation and gender recognition, which aim to automatically determine age and gender attributes based on multimodal cues~\cite{wu2023erdbf, manojlovska2025interpreting}.
Multimodal age estimation is typically formulated as either a regression task (predicting a scalar value)~\cite{pan2018mean} or a classification task (predicting age groups)~\cite{levi2015age}. Multimodal gender recognition is generally treated as a binary classification task~\cite{duan2018hybrid}, distinguishing between male and female.
Compared to vision-only methods, multimodal approaches can capture richer and more synergistic cues, thereby improving the robustness and accuracy of facial soft biometrics recognition.

\subsection{Multimodal Multi-task Joint Learning} 
Facial attributes are inherently interdependent, exhibiting strong anatomical and semantic correlations. For example, a “Happy” expression in FER is anatomically composed of specific AUs (AU6 + AU12), while the manifestation of these AUs can be further modulated by soft-biometric factors such as age.
These intrinsic relationships naturally motivate multi-task joint learning. Specifically, this strategy integrates AU recognition, FER, soft biometrics recognition, and other related tasks to mutually enhance performance within a unified framework~\cite{shao2018deep}.
Furthermore, in multimodal settings, joint learning extends seamlessly to the synergistic fusion of heterogeneous data streams to capture complementary aspects of facial behavior~\cite{kollias2023multi}.
By simultaneously aligning cross-modal representations and modeling the dependencies among multiple attribute labels, the model learns a unified embedding that benefits from both modal cooperation and inter-task relationships. This results in more robust and stable performance, particularly under challenging conditions where single-modal or single-task models often fail.

\section{Datasets}

Traditionally, facial state analysis has been studied under single-modality, single-task settings, with dedicated datasets for each task, such as BP4D~\cite{zhang2014bp4d} for action unit (AU) recognition, AffectNet~\cite{mollahosseini2017affectnet} for facial expression recognition (FER), and UTKFace~\cite{UTKFace} for soft biometric recognition.
More recently, multimodal facial state analysis datasets~\cite{MER2023} have attracted increasing attention due to the enhanced robustness of multimodal signals in real-world scenarios.
Meanwhile, the rapid emergence of general-purpose models has further stimulated interest in multi-task learning, leading to the development of multi-task multimodal datasets~\cite{FaceBench}.
Accordingly, we categorize existing facial state datasets into \emph{single-task} and \emph{multi-task multimodal datasets}, as summaried in Table \ref{tab:dataset_overview2}.

\subsection{Single-task Multimodal Facial State Datasets}
Single-task multimodal facial state datasets~\cite{tawara2021age} remain centered on a single primary task, while extending visual-only inputs to multimodal data.
One line of work incorporates multimodal signals directly during dataset construction.
For instance, DEAP~\cite{koelstra2011deap} dataset employs music videos to elicit affective responses from participants and simultaneously records facial videos together with physiological signals during the stimulation process, providing reliable ground-truth annotations for the evaluation of emotion recognition algorithms.
Another line of research focuses on extending existing vision-only datasets by augmenting them with additional modalities in a post-hoc manner. For example, 
BP4D+~\cite{zhang2016multimodal} extends the original vision-only BP4D~\cite{zhang2014bp4d} dataset, where the original 2D/3D facial recordings and AU annotations are retained, while additional multimodal signals, including physiological measurements and thermal data, are synchronously collected.

\subsection{Multi-task Multimodal Facial State Datasets}
Beyond single-task settings, a growing body of work has focused on multi-task multimodal datasets to facilitate multidimensional facial state analysis.
These datasets are broadly categorized into \emph{structured-label datasets} and \emph{language-supervised datasets}.

The former~\cite{kollias2019expression, li2020eeg} are primarily designed for conventional discriminative models, providing multimodal inputs along with explicit multi-task labels. A representative dataset is Aff-Wild2~\cite{kollias2019expression}, which comprises video and audio data with frame-level annotations of facial AUs, discrete emotion categories, and continuous valence–arousal (VA) values.
More recently, the rapid advancement of multimodal large language models (MLLMs) has encouraged researchers to fine-tune these models for facial analysis, capitalizing on their strong pre-training and generalization capabilities. This paradigm shift has spurred the development of language-supervised datasets~\cite{face-llava, FABA, FaceBench}, which move away from label-centric annotations toward flexible language-based pairs (e.g., visual question–answer pairs and instruction–response pairs).
By replacing categorical labels with open-ended textual responses, these datasets provide more fine-grained supervision that often includes reasoning or descriptive explanations, significantly improving the interpretability of facial analysis tasks.

\textbf{Discussion.} Existing multimodal multitask facial analysis datasets support initial joint modeling of facial AUs, expressions, and soft biometric traits beyond single-modality and single-task settings. However, they remain constrained by inconsistent annotations, limited modality and demographic coverage, weak real-world robustness, and insufficient privacy safeguards. Future work should focus on standardized annotations, enhanced diversity and robustness, explicit task-level semantic integration, and stronger privacy governance to better support multi-task learning and real-world applications.

\begin{figure}[t]
    \centering
		\includegraphics[width=1.0\linewidth]{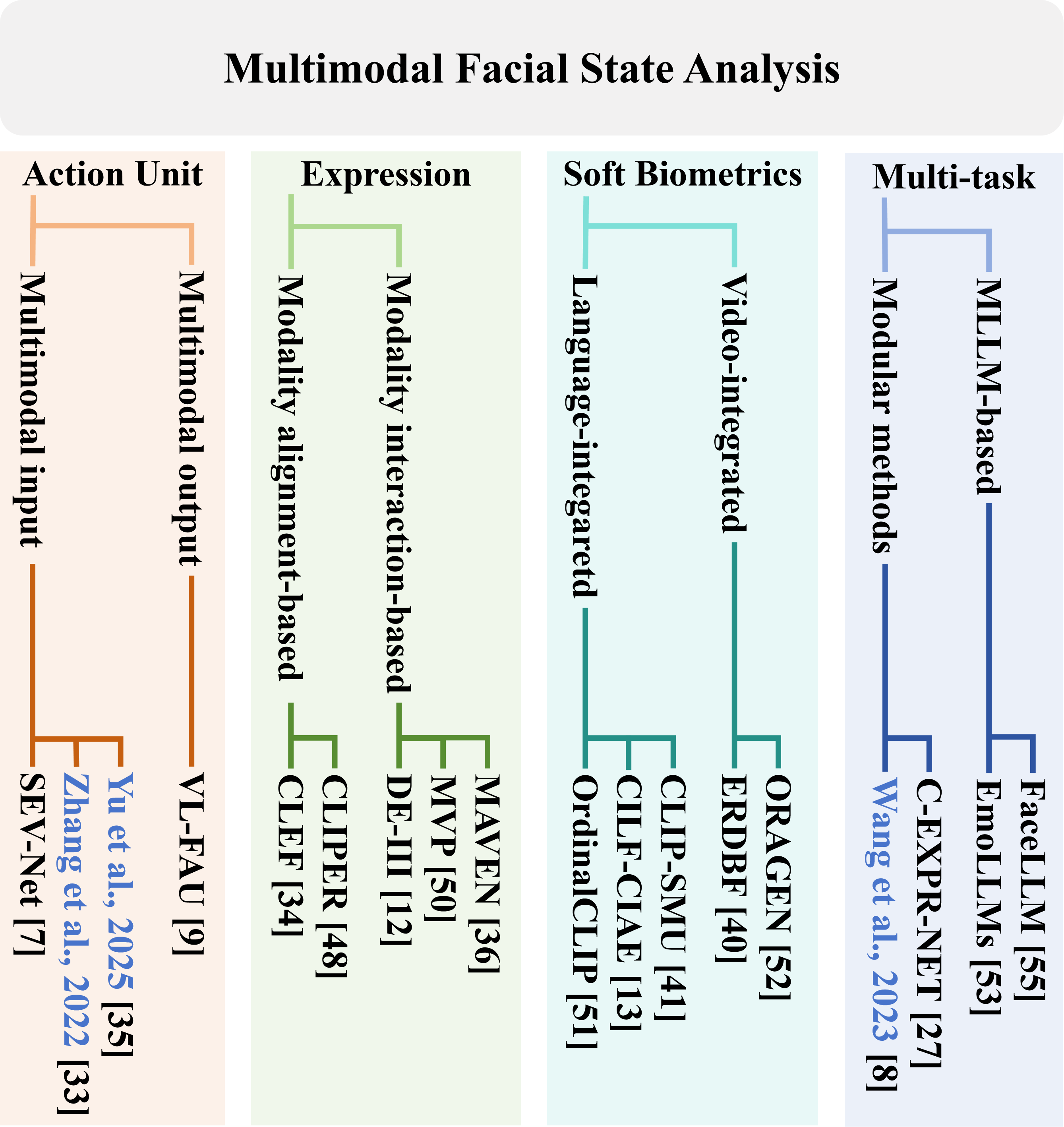}
        \vspace{-1.5em}
     \caption{Taxonomy of multimodal facial state analysis, covering three core tasks and multi-task joint learning, along with representative methods for each category.
     }
    	\label{fig:methods}    
     	\vspace{-1.5em}
\end{figure}

\section{Methods} 
In this section, we review representative multimodal approaches for facial state analysis, covering three major directions: action unit (AU), expression, and soft biometrics recognition. Then we further discuss recent efforts that jointly model these tasks in a multi-task learning framework. Fig.~\ref{fig:methods} illustrates the taxonomy of these approaches along with representative methods for each category.

\subsection{Multimodal Facial Action Unit Recognition}
Vision-based unimodal approaches to facial AU recognition are inherently sensitive to adverse imaging conditions, such as changes in lighting and partial occlusions, which significantly degrade their robustness. Multimodal AU recognition mitigates these limitations by leveraging complementary cues from various modalities, leading to more reliable AU prediction.

The most common multimodal methods input visual data and one or more auxiliary modalities, allowing cross-modal interactions to enhance the robustness of AU recognition.
For example, \cite{yang2021exploiting} leverage textual descriptions of AUs via cross-modal attention to guide visual features toward semantically relevant facial regions, thereby enhancing discriminability. \cite{yu2025towards} employ dual-stream architectures that process video and audio signals separately before fusing modality-specific representations to improve robustness under challenging conditions. \cite{zhang2022transformer} further broadens this research direction by increasing the number of modalities, jointly utilizing visual, audio, and textual information in a tri-branch framework for AU recognition.

Despite their effectiveness, these methods remain black-box systems that offer limited insight into their decision-making processes, thereby restricting their deployment in high-risk domains.
To address this limitation, recent works~\cite{ge2024towards, manojlovska2025interpreting} have proposed generating explanatory text alongside standard AU predictions.
For instance, \cite{ge2024towards} employs language generation modules to produce local and global textual explanations concurrently with AU predictions. By introducing a textual modality at the output stage to supervise visual feature learning, this design not only boosts AU recognition performance but also significantly enhances model interpretability.

\textbf{Discussion.} 
Existing multimodal methods improve robustness by incorporating auxiliary modalities at the input or output level. However, these methods tend to rely heavily on cross-modal integration, while overlooking the structured dependencies among AUs that have proven effective in vision-based approaches. A promising research direction is to jointly consider multimodal cues and inter-AU relationships, enabling more coherent and discriminative AU representations.

\subsection{Multimodal Facial Expression Recognition}
A core challenge in multimodal facial expression recognition (FER) is integrating heterogeneous signals across modalities. Existing methods can be broadly grouped into two paradigms: \emph{modality alignment} and \emph{modality interaction}.

Modality alignment~\cite{zhang2023weakly, li2024cliper} aims to project diverse modalities, such as visual, auditory, and textual signals, into a shared semantic space.
This is often achieved through contrastive learning, which encourages semantically corresponding samples from different modalities to have similar representations.  For instance, \cite{zhang2023weakly} uses weakly supervised contrastive learning to adapt CLIP~\cite{radford2021learning} for expression-aware cross-modal alignment. 
By mapping heterogeneous signals into a shared latent space, modality-alignment methods reduce modality gaps and provide consistent cross-modal supervision, enabling the model to more effectively exploit complementary cues for improved FER performance.

Modality interaction~\cite{wang2023pose, shi2024detail} explicitly models cross-modal exchanges, allowing features from different modalities to be progressively integrated and ultimately fused into a unified representation.
Early multi-branch fusion designs~\cite{strizhkova2024mvp} fuse modality-specific features after independent encoding, while recent methods introduce explicit interactions such as bidirectional cross-attention for audio--visual integration~\cite{shi2024detail}.
This paradigm is particularly effective for modeling subtle or context-dependent expressions, where cues in one modality may modulate or disambiguate signals in another.

\textbf{Discussion.} 
Overall, both modality alignment and modality interaction effectively integrate cross-modal cues, but often depend on heavy Transformers and attention modules, incurring substantial parameter and compute overhead that hinders real-time or resource-constrained deployment; future work should prioritize lightweight multimodal designs that preserve cross-modal representation power while improving efficiency.

\subsection{Multimodal Facial Soft Biometrics Recognition}
Recent advances in multimodal learning have significantly advanced methodological developments in facial soft biometrics recognition, particularly in age estimation and gender recognition. Among existing approaches, the modalities most commonly integrated with visual information are language and audio.
The incorporation of language modality aims to inject the high-level, rich semantics from textual descriptions into facial visual features. Prior works~\cite{li2022ordinalclip, shou2023cilf} typically build on pretrained vision–language backbones such as CLIP~\cite{radford2021learning}, where images and attribute-related texts (e.g., “a 30-year-old face,” “female with youthful appearance”) are projected into a shared embedding space.
For example, \cite{manojlovska2025interpreting} employs images paired with template textual descriptions to jointly infer age, gender, and other attributes based on feature similarity within a shared multimodal embedding space.
These approaches benefit from the inherent linguistic structure captured by large language models, which provides contextual priors and reduces the reliance on large-scale annotated facial datasets.

Age and gender influence not only facial appearance but also vocal characteristics, such as pitch, timbre, and speaking style. Inspired by this intrinsic correlation, acoustic information is frequently incorporated as an auxiliary modality to enhance visual cues~\cite{wu2023erdbf, markitantov2025audio}, thereby improving robustness when image quality is degraded or faces are occluded.
These approaches typically adopt a dual-branch architecture, encoding visual and audio data separately, which are then integrated through a fusion module. Audio-enhanced age estimation leverages biologically correlated speech patterns to refine predictions, while gender recognition benefits from characteristic differences in vocal frequency distributions.

\textbf{Discussion.}
Multimodal methods enhance facial soft biometrics recognition by leveraging complementary cues, but most rely on fixed modality combinations and degrade when modalities are missing or corrupted. Future work should develop robust strategies that handle absent or incomplete modalities to improve real-world adaptability and generalization.

\subsection{Multimodal Multi-task Joint Learning}
Multimodal multi-task joint learning aims to integrate heterogeneous data sources while modeling the inherent relationships among facial analysis tasks.
Conventional modular approaches~\cite{wang2023pose, kollias2023multi} typically employ modality-specific encoders to extract features from different data sources, followed by fusion modules, such as cross-attention, to build a shared multimodal representation. This unified representation effectively integrates information across modalities.
Lightweight task-specific heads (e.g., fully connected layers) are then added on top of the shared embedding to generate AU, expression, or soft-biometric predictions, enabling the framework to balance shared feature learning with task-level specialization.

Recent advances in multimodal large language models (MLLMs) have introduced a new paradigm for this task. A single MLLM, equipped with domain-oriented tuning~\cite{liu2024emollms, li2025faceinsight}, can jointly handle diverse facial analysis tasks without relying on manually designed branches. For example, \cite{shahreza2025facellm} fine-tunes MLLMs using high-quality facial question-answer pairs, achieving strong performance across a wide range of face-centric tasks.
Beyond delivering unified predictions, MLLMs further produce fine-grained natural-language explanations of their predictions by leveraging their strong abilities in multimodal perception, understanding, and reasoning. This provides, to a certain extent, valuable interpretability and transparency that traditional discriminative models rarely exhibit.

\textbf{Discussion.}
Early modular methods rely on carefully designed cross-modal fusion mechanisms and task-specific interactions, which can be complex and labor-intensive. In contrast, MLLM-based approaches heavily depend on the foundational abilities of the MLLM and typically require substantial computational resources for fine-tuning. Despite these challenges, we believe that multimodal multi-task paradigm remains highly promising.
Future work could continue to explore this paradigm by developing more sophisticated model architectures and more efficient fine-tuning strategies, thereby further enhancing performance in facial analysis tasks.

\section{Conclusion}
This survey reviews multimodal facial state analysis across AU recognition, FER, facial soft biometrics, and multimodal multi-task learning. By fusing visual, audio, textual, physiological, and other complementary modalities, multimodal approaches overcome the robustness and interpretability limits of unimodal methods, while multi-task learning exploits intrinsic anatomical and semantic correlations to improve fine-grained modeling and generalization. Despite progress in modality alignment, interaction, and language-integrated paradigms, challenges remain in annotation consistency, demographic and modality coverage, computational efficiency, and robustness to missing modalities. Future work should focus on standardized datasets, lightweight and robust multimodal fusion, and deeper task-level integration to enable practical and interpretable facial state analysis systems.

\bibliographystyle{IEEEbib}
\bibliography{icme2025references}
\end{document}